\documentclass[manuscript,nonacm]{acmart}
\usepackage[utf8]{inputenc}
\usepackage{tikz}
\usepackage{standalone}
\usepackage{tikz-3dplot}
\usepackage{subfig}
\usepackage{graphicx}
\usepackage{wrapfig}
\usepackage{enumerate}
\usepackage{pgfplots}
\pgfplotsset{compat=1.17} 

\usetikzlibrary{backgrounds,scopes,fit,positioning}
\definecolor{my_red}{HTML}{FF0F19}
\definecolor{my_grey}{RGB}{128,128,128}
\definecolor{my_green}{RGB}{1,68,33}
\definecolor{my_yellow}{HTML}{FDE9CA}
\definecolor{my_highlight}{HTML}{E99076}
\definecolor{my_gray}{HTML}{A5A5A5}
\newcommand{\Until}{\mbox{$\, {\sf U}\,$}}

\title{Dynamic Certification for Autonomous Systems}

\begin{document}

\author{Georgios Bakirtzis}
\affiliation{%
   \institution{The University of Texas at Austin}
   \city{Austin}
   \state{TX}
   \country{USA}}
\email{bakirtzis@utexas.edu}

\author{Steven Carr}
\affiliation{%
   \institution{The University of Texas at Austin}
   \city{Austin}
   \state{TX}
   \country{USA}}
\email{stevencarr@utexas.edu}

\author{David Danks}
\affiliation{%
   \institution{University of California, San Diego}
   \city{Austin}
   \state{TX}
   \country{USA}}
\email{ddanks@ucsd.edu}

\author{Ufuk Topcu}
\affiliation{%
   \institution{The University of Texas at Austin}
   \city{Austin}
   \state{TX}
   \country{USA}}
\email{utopcu@utexas.edu}
\maketitle

Autonomous systems are often deployed in complex sociotechnical environments, such as public roads, where they must behave safely and securely. Unlike many traditionally engineered systems, autonomous systems are expected to behave predictably in varying ``open world'' environmental contexts that cannot be fully specified formally. As a result, assurance about autonomous systems requires us to develop new \emph{certification methods}---codified checks and balances, including regulatory requirements, for deploying systems---and mathematical tools that can dynamically bind the uncertainty engendered by these diverse deployment scenarios.
More specifically, autonomous systems increasingly use algorithms trained from data to predict and control behavior in previously unencountered contexts. Using \emph{learning} is a critical step to engineer autonomy that can successfully operate in \emph{heterogeneous contexts}, but current certification methods need to be revised to address the dynamic, adaptive nature of learning. The heterogeneity that any certification framework ought to address for the design of autonomous systems is twofold. The first relates to the system itself and the heterogeneous components that engender its behavior. The second is the heterogeneity that certifying must address in relation to the complex sociotechnical settings in which the system is expected to behave.%

We propose the \emph{dynamic certification} of autonomous systems---the iterative revision of permissible $\langle \text{use}, \text{context} \rangle$ pairs for a system---rather than prespecified tests that a system must pass to be certified. Dynamic certification offers the ability to ``learn while certifying,'' thereby opening additional opportunities to shape the development of autonomous technology. This type of comprehensive, exploratory testing, shaped by insights from deployment, can enable iterative selection of appropriate contexts of use. More specifically, we propose dynamic certification and modeling involving three \emph{testing stages}: early-phase testing, transitional testing, and confirmatory testing. Movement between testing stages is not unidirectional; we can shift in any direction depending on our current state of knowledge and intended deployments. We describe these stages in more detail below, but the key is that these stages enable system designers and regulators to learn about and ensure that autonomous systems operate within the bounds of acceptable risk.

Our proposal is similar to how the Food \& Drug Administration (\textsc{fda}) tests drugs in stages with increasing scrutiny before being approved for public consumption. Rather than a simple yes/no certification, the \textsc{fda} uses an iterative process of exploratory stages in which pharmaceutical agents are first approved for limited uses in restricted contexts under careful oversight and only gradually approved for broader uses as post-approval monitoring and subsequent studies demonstrate safety and efficacy. Of course, the \textsc{fda} procedures cannot be used directly for dynamic certification of autonomous (software) systems, but they provide an ``existence proof'' that dynamic certification can work.

Technology creation involves at least two different yet interdependent types of decisions. \emph{Design decisions}  determine the structure and intended operation of the autonomous system, including the evaluation functions that are optimized during development and revision/updates. \emph{Deployment decisions} determine the contexts and uses for the autonomous system, including designating certain situations as ``do not use'' (or ``use only with increased oversight''). In practice, static certification and regulatory systems often focus only on deployment decisions (and take the design decisions and technical specifications as fixed). However, precisely because of the frequent uncertainty about what counts as ``success'' for an autonomous system, certification of those systems must also consider design decisions, using technical specifications to predict performance in unencountered contexts. %

Dynamic certification includes design decisions, particularly in the early stages when changes have the highest impact and lowest cost, often before code or hardware have even been built~\cite{frola:1984,strafaci:2008}. Mathematical tools from formal methods can thus play an essential role in specifying autonomous systems at different levels of abstraction, even when they have not yet been implemented. Formal methods allow us to specify acceptable risks, identify failures that inform mitigation strategies, and understand and represent the uncertainty associated with deploying autonomous systems in heterogeneous environments. Formal models are also living documents that encode design and deployment decisions made throughout the lifecycle of the autonomous system. For example, tracking changes in the specification of requirements throughout the lifecycle can give a good picture of the design problems and solutions at a particular time and how those changes reflect design shifts over time. Successful dynamic certification thus depends on translational research by formal methods, autonomous systems, and robotics communities to establish proper procedures to ensure that deployed systems are unlikely to cause harm. 

Dynamic certification relies on an iterative assessment of the risks (and benefits) introduced by deploying autonomous systems for different uses and contexts. Formal methods offer a concrete basis for specification, verification, and synthesis for autonomous systems but do not guide the translation of our desired values and acceptable risk into those formal models. We require frameworks that explicitly allow for ambiguities in specifications and uncertainties and partial decisions in modeling while remaining scalable to practically relevant sizes. More generally, dynamic certification will require an appropriate coevolution of regulatory and formal frameworks. Having argued for implementing parts of dynamic certification via formal methods, it is crucial to acknowledge other types of analyses that could implement dynamic certification, such as assurance cases~\cite{asaadi:2020}, structured interrogation of requirements~\cite{leveson:1994,webster:2020,bourbouh:2021}, and domain standards~\cite{farrell:2021}. Indeed, these other types of methods and their associated tools and metrics could play valuable roles in the dynamic certification of autonomous systems.

\begin{figure}[!t]
    \centering
    \subfloat[Suburban context]{\includegraphics[width=0.48\textwidth]{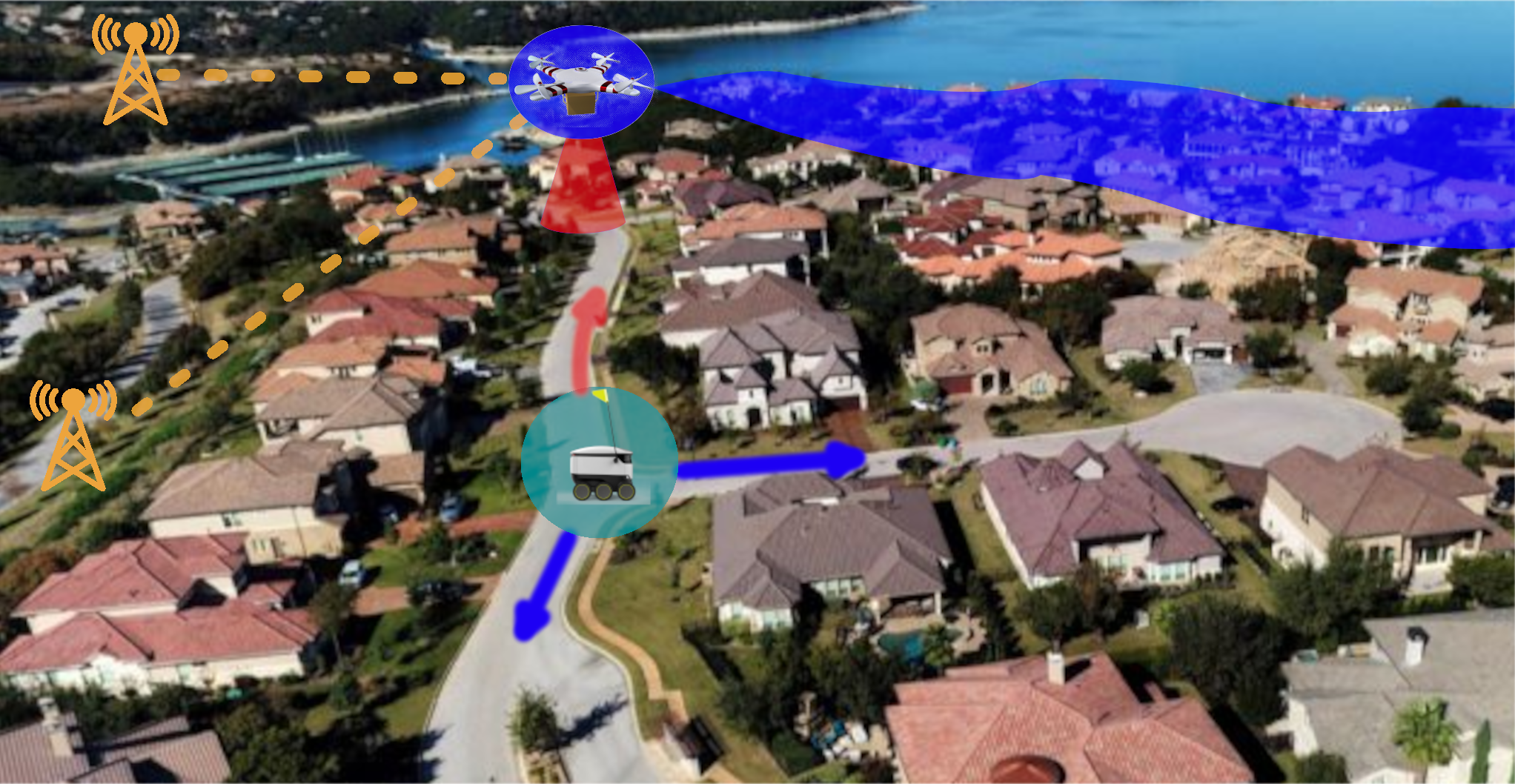}}\hfill
    \subfloat[Urban context\label{fig:scenario-urban}]{\includegraphics[width=0.48\textwidth]{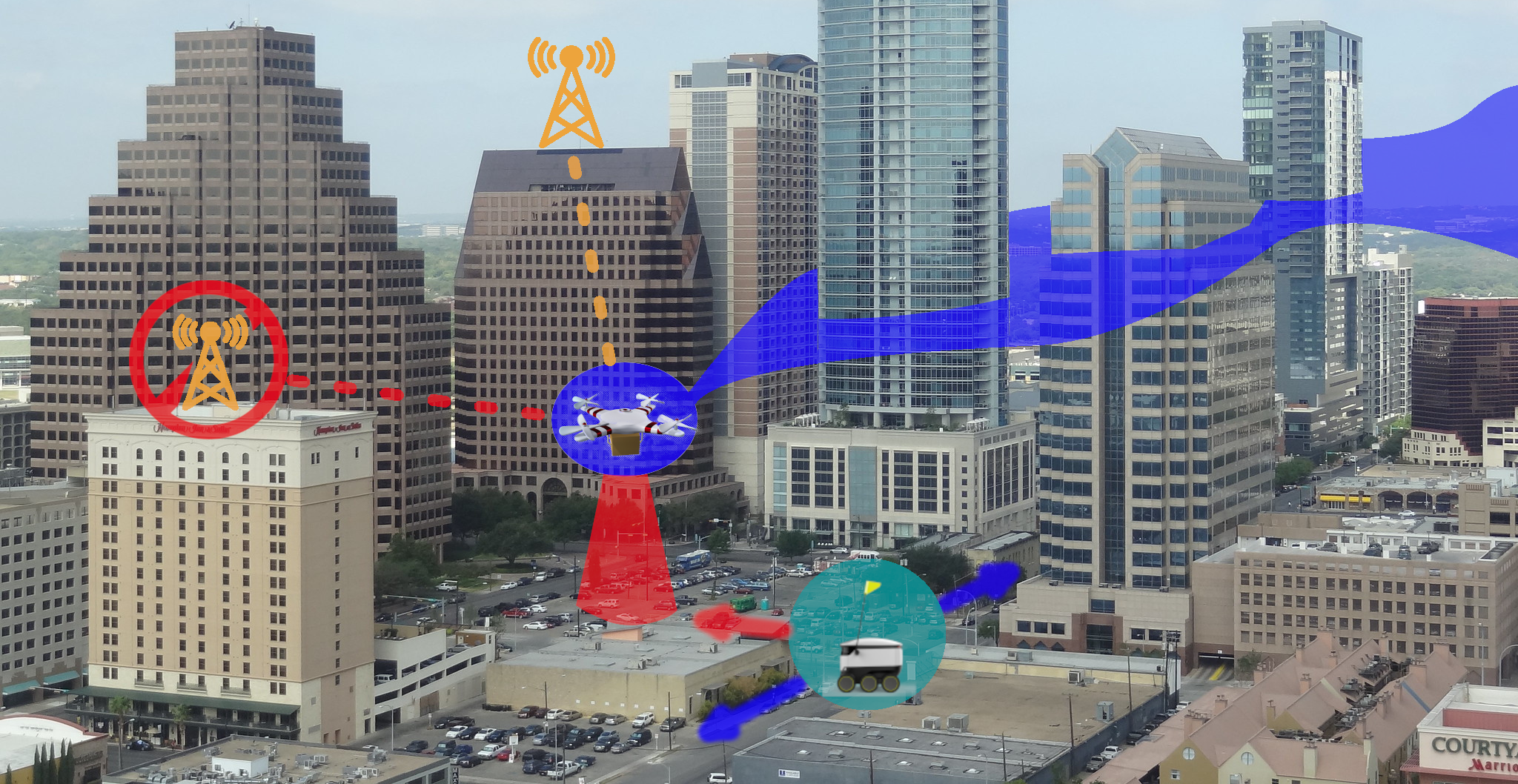}}\vspace{-1em}
    \caption{Different $\langle \text{use}, \text{context} \rangle$ pairs result in different hazard conditions. The arrows and shaded regions are possible trajectories, safe (blue) and hazardous (red). Dotted lines indicate connectivity. (a) In suburban areas, the probability of losing connection is low because the signal has minimal attenuation. (b) In a city, buildings are made with concrete and rebar and are laid out densely, increasing the probability of signal attenuation. Therefore, the system is more likely to transition to a hazardous state in the city $\langle \text{use}, \text{context} \rangle$ pair. When attempting to deploy in these different scenarios we need to check different safety properties. Dynamic certification requires the development of a framework for designers that easily adapts between the different scenarios.}
    \label{fig:scenario}
\end{figure}

\vspace{.5em}
\noindent\textbf{Scenario.} \quad
We motivate and illustrate our proposed framework for dynamic certification using a scenario with two interacting systems,
an unmanned aerial vehicle (\textsc{uav}) and a ground-based delivery robot simultaneously delivering packages~(Fig.~\ref{fig:scenario}). We require that the \textsc{uav} only operate while connected to a wireless communication network. If it (probabilistically) loses connection, it must land in place. A hazardous state results if the \textsc{uav} lands in the same location as the ground-based robot. The designer for the \textsc{uav} seeks a high-level risk-mitigation strategy that accounts for the ground-based robot's movement and limits the probability of transitioning to a hazardous state. This strategy requires specification of acceptable $\langle \text{use}, \text{context} \rangle$ pairs, e.g., usable trajectories or locations for the \textsc{uav}. This high-level focus enables key abstractions and simplifications. For instance, the designer can abstract away the low-level \textsc{uav} controller and assume that it can safely navigate between waypoints. Instead of modeling complex behaviors about the ground-based robot---which may not even be possible if the system is outside of the designer's control---the designer can require that the \textsc{uav} be robust against random movements of the robot. Though this example is simplified in various ways (e.g., only including two robots), those simplifications serve to highlight key conceptual points, including the value of formal methods.

 \section*{Dynamic certification}

Dynamic certification is built on two fundamental operations, modeling, and testing. Modeling allows system engineers to keep track of design choices that otherwise would be difficult to document and adjust when issues arise because of complex interactions between subcomponents. Models also enable the engineer to focus on the interfaces between subcomponents, often abstracting away the individual subcomponents to focus on the behavior of the whole. Importantly, we can use models to understand how the system might succeed or fail before the system is built---that is, for high-impact, low-cost design decisions. In contrast, testing involves the actual implementations, focusing on whether the assumptions of the model and the resulting design decisions actually function as expected in the physical world. Conventional static certification struggles when operational (or regulatory) assumptions fail to hold in reality. In contrast, dynamic certification posits that modeling and testing should be intertwined throughout the system lifecycle, so our models (and assumptions) can be continually refined as we better understand real-world contexts. Certification of autonomous learning systems requires both elements: testing since the world can surprise us and the system can change through learning, plus modeling to guide our design and testing decisions through the massive search spaces. 

Assurance requires specifying when, where, and why an autonomous system is being deployed within a sociotechnical context. But if autonomous systems are expected to learn from their environment and context of operation, then there does not seem to be a stable model for testing. Dynamic certification turns this concern into a virtue: if our base system model contains appropriate parameters, we can iteratively refine and augment this base model through different testing procedures. This virtue and the resulting testing procedures come from the feedback and interaction between stakeholders with different concerns and expertise, making it clear when testing procedures are sufficient and accurate. Therefore, in the long run, we can conduct sufficient testing to have an accurate model that assures stakeholders that systems will operate as expected. 

Specification of the base system model for dynamic certification requires (perhaps partial) identification and description of the following four components (inspired by Kimmelman and London \cite{kimmelman:2015}).
\begin{itemize}
    \item \textsc{Modules} of the system (primarily software, but potentially hardware) including the function(s) of each module.
    \item \textsc{Contexts} in which the system is expected to be capable of successful operation.
    \item \textsc{Mappings} from Context $\rightarrow$ Behavior for ``successful'' performance in various conditions.
    \item \textsc{Variations} in the environment for which the system should be robust.
\end{itemize}

Given an initial specification of these four elements for an autonomous system, dynamic certification can be divided into three distinct stages (with no requirement for unidirectional progression through these stages). All four components of the base model specification can be revised or adjusted during each stage. Although discussions of certification often focus on \textsc{contexts} and \textsc{mappings}, the inclusion of design decisions in dynamic certification means that other components can also be adjusted (e.g., adding \textsc{modules} to improve performance in given \textsc{contexts}).

The first stage is \emph{early-phase testing}, which occurs in the development lab or other highly controlled settings. The two main goals of this stage are (1) to verify that the integrated \textsc{modules} implement the intended \textsc{mappings}; and (2) to develop appropriate base models of the autonomous system for offline testing. The first goal is relatively standard when developing a software system (for example, unit-testing). The second goal, however, is much less common and requires careful consideration of the range of \textsc{contexts} and \textsc{variations} that might be encountered in plausible deployment environments. Importantly, all four components of the base system model must be (tentatively) specified in early-phase testing; this stage is \emph{not} solely technology-focused. Given an initial specification, early-phase testing continues until the software system is suitably verified and its expected performance is sufficiently good in offline testing. In the running scenario, early-phase testing could take the form of building and testing a gridworld that models the high-level decision-making for the \textsc{uav}. In this stage, the designer would identify anomalous behavior, such as locations that create deadlocks, thereby enabling design decisions to mitigate situations that lead to task degradation~\cite{fleming:2021}.

The second stage is \emph{transitional testing} in which the system is deployed in real-world environments, though with significant oversight and control. The two main goals of this stage are (1) to identify \textsc{contexts} of real-world failure; and (2) to characterize potential environmental \textsc{variations}. These goals require highly active engagement and interventions; this stage is not simply ``deploy and watch'' or ``compare to prior standards.'' Rather, transitional testing should involve, for example, focused efforts to place the system into ``hard'' contexts precisely to improve our understanding of the system. Transitional testing involves careful, systematic efforts to determine the boundaries of appropriate system performance. The information produced by this testing can be iteratively used to change \textsc{modules}, constrain \textsc{contexts}, add \textsc{mapping} complexity, or increase \textsc{variation} specificity. Transitional testing is exploratory (helping to understand), not merely confirmatory (checking if the system performs as expected). In our running scenario, transitional testing would involve \emph{testing} (not just modeling) system performance with high-fidelity and hardware-in-the-loop simulations~\cite{bacic:2005,shah:2017,curiel:2019} or in controlled environments (e.g., a large industrial park with limited public traffic). This stage intends to gather enough data to modify the formal system model to reflect reality further. 

The third and final stage is \emph{confirmatory testing} in which the system is deployed with significant oversight and monitoring, but no further controls beyond those specified in the certification by a set of $\langle \text{use}, \text{context} \rangle$ pairs. This stage aims to determine, in real-world settings, both (1) system performance reliability; and (2) the extent of system-user value (mis)matches. The latter goal is crucial because many autonomous system ``failures'' involve a properly functioning system that implements different values than the users expect. The system behaves correctly, but according to a (perhaps implicit) notion of ``success'' that is different from that of the human users;\footnote{Many classic examples of ``AI run amok'' fall into this category. For example, the paperclip maximizer \cite{bostrom:2006} simply has a different idea of ``success'' than us.} that is, the system implements the wrong \textsc{mapping}. These divergences often appear only once the system is in the hands of untrained users, so confirmatory testing must initially include significant oversight to detect, record, and respond to real-world performance failures and value divergences. This monitoring can be gradually reduced as we learn the exact behavior of the system in relevant real-world contexts (i.e., even this stage involves some exploratory testing).\footnote{Confirmatory testing is thus quite similar to conformance testing but does not assume that we have a fully-specified set of standards and behaviors that are provided in advance.} In the running scenario, confirmatory testing would involve supervised deployment in a controlled environment, possibly borrowing rules and regulations from the operational design domain~\cite{koopman:2019}. Changes to the system design based on actual operational contexts should reflect the formal model; they must agree. Once testing and modeling agree, the dynamic certification has ensured that the system will behave acceptably and safely.

Current static certification frameworks involve testing only late in the lifecycle after a particular system implementation has been built and is often already deployed. They could theoretically play a role beyond setting performance targets, but in practice, they rarely do. In contrast, dynamic certification uses testing throughout the lifecycle, revealing challenges and tradeoffs while design decisions and changes are still possible. The benefits of lifecycle-wide testing require models that can capture the \emph{what's}, \emph{why's}, and \emph{how's}, along with connections to the eventual system design. %
Formal models play a particularly valuable role in dynamic certification. In particular, formal models can be used early to interrogate our assumptions about the system's requirements rather than only being used late to provide provable guarantees. Formal methods also can give us the tools to add stakeholder-specific semantics to various models of behaviors, requirements, and architectures, thereby providing a common language to reason about the system's design. 

\section*{Formal methods for dynamic certification}

Formal models use the precision of mathematical language to reveal misunderstandings about the system's behavior and requirements~\cite{wing:1990,lamport:2002,fisher:2013,lukcuck:2019}. Formal specifications can model complex systems before developing code or synthesizing hardware architectures, allowing systems engineers to interrogate requirements and find clashes and interaction faults early in the system's lifecycle. Using formal models, we can \emph{architect} a system proactively: no system exists yet, so our design decision effectiveness is highest and the cost of changes lowest since we do not have to bolt modifications onto a preexisting design.
Additionally, we can often \emph{synthesize} behaviors directly from the formal model, which then provides our implementation with guarantees about properties we care about, such as safety~\cite{seshia:2015}. Finally, formal models can \emph{inform} testing procedures by simulating different contexts and becoming more comprehensive (and therefore informative) as system data are collected during deployment~\cite{kress:2021,fan:2019}---with the caveat that there will always be a need to interpret those formal results to account for the gap between formal models and reality. It is impossible to make autonomous systems 100\% safe 100\% of the time, but we posit that formal methods can significantly assist in designing better, safer systems.

In particular, formal methods can be highly valuable for the dynamic certification of an autonomous system. Formal models can specify behavior and system dynamics that are difficult to implement and test without committing to a specific design. Formal models can therefore be used as an aid to inform what testing ought to take place to ensure that the system will behave as expected. In addition, formal models and specifications can readily be updated given new data to achieve increasing analysis precision as the system is deployed.

One formal model for autonomous systems that is especially useful for dynamic certification is the Markov decision process (MDP). MDPs model sequential decision-making in stochastic systems with nondeterministic choices~\cite{puterman:2014}. They have been useful for modeling high-level decisions in autonomous systems, such as collision-avoidance~\cite{temizer:2010}, surveillance using ground-based robots~\cite{lahijanian:2012}, and transmission exchange for wireless sensor networks ~\cite{alsheikh:2015}. Analysis with MDPs typically requires that the complete model be known a priori~\cite{junges:2019}, but there is often significant model uncertainty in early phases since many design and deployment decisions have yet to be made. We can instead use a class of model known as a {\it parametric} MDP~\cite{hahn:2011}, where parameters model variations in transition probabilities. The parameters may thus represent design choices (e.g., requiring a perception \textsc{module} with a certain error rate, or setting specific thresholds for underlying decision-making algorithms); deployment decisions and context characteristics (e.g., possible reductions in visibility or likelihoods of interruption of information flow); or modeling uncertainties (e.g., unknown characteristics of motion or reaction time under off-nominal conditions). Parametric MDPs have the specificity and flexibility required for a base system model that can be refined and improved through exploratory early-phase testing. 

\begin{figure}[!t]
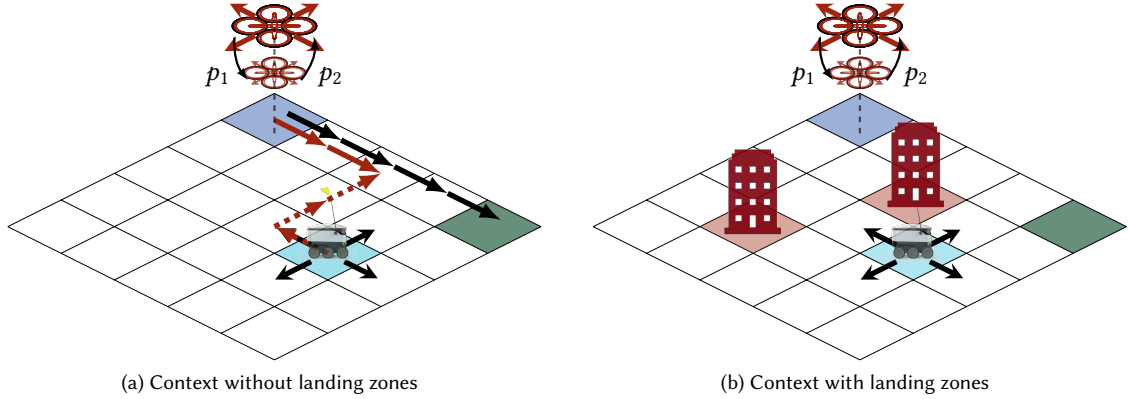

    \centering
    \subfloat[Context without landing zones]{\includestandalone[width=.485\textwidth]{figures/UAV_example}}\hfill
    \subfloat[Context with landing zones\label{fig:ContextLZ}]{\includestandalone[width=.485\textwidth]{figures/UAV_restrict}}
    \caption{Using formal methods we can interrogate models containing different \emph{contexts}. For example,
    a \textsc{uav} (shown in red) deployed in a city (left) is more likely to lose connection and be forced to land (opaque \textsc{uav}) compared to a similar \textsc{uav} operating in a suburban area. For the city context, the value for probability $p_1$ would be larger than the suburban context for the same parameter. For this reason, we ought to account and understand how the same system can be less or more likely to encounter hazardous conditions when interacting with ground-based agents (landing in the vicinity of the delivery robot). We include a possible counterexample whereby the \textsc{uav} drops connection and crash lands on the delivery robot.
     In a similar, albeit slightly modified context (right), we can choose to explicitly eliminate the possibility of ground-based interactions by having the \textsc{uav} enter the ground layer in a prescribed controlled landing zone in the form of the building roofs (shaded red).}
     \label{fig:context}
\end{figure}

We illustrate the use of parametric MDPs as an early-phase decision-making tool in our running scenario with two autonomous systems,
a \textsc{uav} and a ground-based delivery robot, simultaneously delivering packages~(Fig.~\ref{fig:context}).
For dynamic certification, we want to iteratively identify uses and contexts for which the \textsc{uav} can safely deploy while continually gathering additional data to determine when it can be deployed in more heterogeneous environments. Safe deployment is critical in all phases (not just confirmatory testing) due to the possibility of problematic incidents. For example, if the \textsc{uav} were to hit a delivery robot in early testing, then even if there were no damage to either system, this reportable event might delay the \textsc{uav}'s certification and eventual deployment. Mitigating these issues at design time makes it less likely that such an event would occur and more likely that the system would deploy within schedule. 

Suppose that we have a list of formal requirements (perhaps translated from human values) for the \textsc{uav} performance. Parametric MDPs provide a useful model to check with what probability these properties hold or, perhaps more importantly, do not hold.
In this context, the \textsc{uav} computes a policy that maximizes the probability of satisfying a temporal logic specification. Based on a finite number of samples of the uncertain parameters, each of which induces an MDP, we can estimate the best-case probability that the policy satisfies the required specification by solving a finite-dimensional convex optimization problem.

More specifically, we have three relevant, high-level \textsc{modules} (Fig.~\ref{fig:context}) that determine the movement of the \textsc{uav}, the movement of the robot, and the communication status of the \textsc{uav}.
When translating a physical environment such as the predefined scenario (Fig.~\ref{fig:scenario}) into a formal model, we abstract roadway intersections as states in a gridworld (Fig.~\ref{fig:context}). While gridworlds represent rather simplistic modules, they are quite powerful in demonstrating scalable behavior. Simply, an agent that fails to behave safely in such simple environments is
also unlikely to behave safely in real-world~\cite{leike:2017}.
A parametric MDP can model the composition of these three modules into a single sociotechnical system. 
The \textsc{uav} can land and take-off from anywhere in the region. It will lose connection and land-in-place with probability $p_1$ (opaque \textsc{uav} in Fig.~\ref{fig:context}), and remain grounded until it reestablishes connection with probability $p_2$. 
We formalize the \textsc{uav} goal of ``safely deliver the package'' as the requirement that the \textsc{uav} behavior maximizes the probability 
that it delivers a package to the green region while not creating an \emph{incident} 
by landing in the same physical location as the delivery robot.
We describe such a mission using the temporal logic formula $\phi = \neg \textrm{Crash} \Until \textrm{Goal}$, where $\textrm{Crash}$ is true when a landed \textsc{uav} shares the same location as the delivery robot.
We thus abstract away complex low-level interactions involving landing 
or taking-off in a crowded region,
and instead focus on the human-relevant behavioral understanding and characterization of what might go wrong.

\begin{figure}
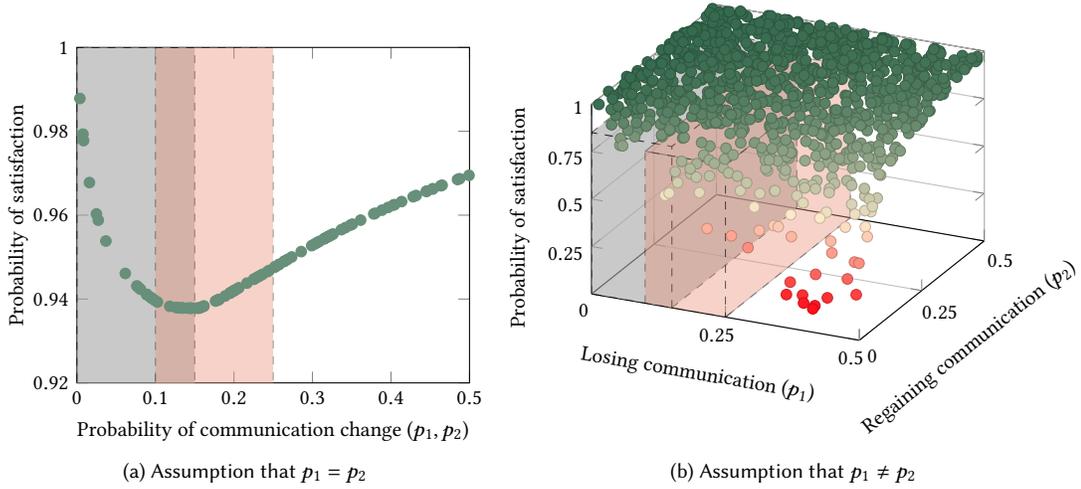

    \centering 
    \subfloat[Assumption that $p_1=p_2$\label{fig:C1res}]{\input{figures/grid_data}

\begin{tikzpicture}
	\begin{axis}[name=context1,
	xlabel = {Probability of communication change ($p_1,p_2$)},
	ylabel = {Probability of satisfaction},%
	ylabel style={yshift=-0.15cm},
	xmin=0,
	xmax=0.5,
 	ymin= 0.92,
	ymax= 1.0,
	label style={font=\small},
	tick label style={font=\small},
	width = 0.45*\textwidth,
	height = 0.4*\textwidth,
	legend cell align={left},
	colormap={ryg}{color=(my_red) color=(my_yellow) color=(my_green)}
	]
    \addplot[my_green!60,only marks] table[x index={0}, y index={1}] from \contextOneLine;
    \addplot[no marks,dashed, fill=my_gray,opacity=0.6] coordinates {(0,0) (0.15,0.0) (0.15,1.0) (0.0,1.0) (0,0)};
    \addplot[no marks,dashed, fill=my_highlight,opacity=0.4] coordinates {(0.1,0) (0.25,0.0) (0.25,1.0) (0.1,1.0) (0.1,0)};
	\end{axis}
\end{tikzpicture}}
    \subfloat[Assumption that $p_1 \neq p_2$ \label{fig:C2res}]{\input{figures/grid_data}

\begin{tikzpicture}
	\begin{axis}[
	xlabel = {Losing communication ($p_1$)},
	ylabel = {Regaining communication ($p_2$)},
	xtick = {0,0.25,0.5},
	xlabel style= {sloped like x axis,xshift=-0.0cm},
	ytick = {0,0.25,0.5},
	ylabel style= {sloped like y axis,xshift=-0.0cm},
	ztick = {0.25,0.5,0.75,1.0},
	zlabel = {Probability of satisfaction},%
	ymin= 0,
	ymax= 0.5,
	xmin=0,
	xmax=0.5,
	zmin= 0,
	zmax=1.0,
	label style={font=\small},
	tick label style={font=\small},
	3d box=background,
	grid = major,
	width = 0.45*\textwidth,
	height = 0.4*\textwidth,
	colormap={ryg}{color=(my_red) color=(my_yellow) color=(my_green!80)}]
	\addplot3[scatter,only marks] table [x index={0}, y index={1}, z index={2}]{\contextTwoGrid};
	\addplot3[no marks,dashed, fill=my_gray,opacity=0.6] coordinates {(0,0,0.85) (0,0.5,0.85) (0.15,0.5,0.85) (0.15,0,0.85) (0,0,0.85)};
	\addplot3[no marks,dashed,fill=my_gray,opacity=0.6] coordinates {(0,0,0.0) (0,0.0,0.85) (0.15,0.0,0.85) (0.15,0,0.0) (0,0,0.0)};
	\addplot3[no marks,dashed,fill=my_gray,opacity=0.6] coordinates {(0.15,0,0.0) (0.15,0.5,0.0) (0.15,0.5,0.85) (0.15,0,0.85) (0.15,0,0.0)};
		\addplot3[no marks, fill=my_highlight,opacity=0.4] coordinates {(0.1,0,0.8) (0.1,0.5,0.8) (0.25,0.5,0.8) (0.25,0,0.8) (0.1,0,0.8)};
	\addplot3[no marks, dashed,fill=my_highlight,opacity=0.4] coordinates {(0.1,0,0.0) (0.1,0.0,0.8) (0.25,0.0,0.8) (0.25,0,0.0) (0.1,0,0.0)};
	\addplot3[no marks, dashed,fill=my_highlight,opacity=0.4] coordinates {(0.25,0,0.0) (0.25,0.5,0.0) (0.25,0.5,0.8) (0.25,0,0.8) (0.25,0,0.0)};
	
	\end{axis}
\end{tikzpicture}}
    \caption{The probability that the \textsc{uav} satisfies the specification when employing its optimal policy for attempting to deliver a package without incident to the green square in the gridworld model (Fig.~\ref{fig:context}). In the highlighted regions, we show the expected parameter range for $p_1$ for both suburban (gray) and urban (pink). In Fig~\ref{fig:C2res} the samples in green are likely to complete the delivery without incident while the \textsc{uav}s operating in the environments shown in yellow and red samples are more likely to land-in-place increasing the probability of an incident occuring.}
    \label{fig:Results}
\end{figure}

For the range of parameter values, 
we compute policies for the system using the Storm probabilistic model checking tool~\cite{dehnert:2017}.
When synthesizing the optimal policy, i.e.~the policy that satisfies the expression $p_{max}\left[\phi\right]$, we can also compute the probability that an agent employing this policy will satisfy this mission (Fig.~\ref{fig:Results}).
These probabilities can then be used to provide crucial guidance in the dynamic certification process. 

For instance, when beginning the early-phase testing stage, the designer has minimal insight into the values of $p_1$ or $p_2$. 
One possible outcome is that initially, the designer may assume that these probabilities correlate with signal strength and are, therefore, equal; that is, $p_1=p_2$. In such a case, the formal model is a parametric MDP with a single parameter.
Under this assumption, we can certify that the agent will successfully perform its mission no worse than $\sim$93\% of the time (Fig.~\ref{fig:C1res}).
However, during transitional or confirmatory testing, we may gather more information about the system and learn that $p_1 \neq p_2$.
In light of this new information, we can return to early-phase testing to reconsider the \textsc{uav} behavior (in this environment) as modeled by a parametric MDP with two parameters. 
In the process of synthesizing these policies, we can now compute the probabilities of success across values
for both parameters (Fig.~\ref{fig:C2res}).

The integrated modeling and testing in dynamic certification can lead us to specify a threshold on $p_1$ or $p_2$ for safe deployment. We might identify specific, measurable features that define appropriate deployment contexts. For example, we might require that $p_1 \in [0,0.15]$ (highlighted gray in Fig.~\ref{fig:Results}). Our current design in suburban contexts might satisfy this constraint but require additional changes for urban contexts. We might adjust the design of the \textsc{uav} (e.g., using a more reliable communications device) or instead adjust the context (e.g., providing additional signal towers). In either case, we can justifiably determine the systems, uses, and contexts where safe deployment can be assured (to a given probability).

Alternately, an urban context such as  Fig.~\ref{fig:scenario-urban} could include buildings that provide safe landing zones for the \textsc{uav} (Fig.~\ref{fig:ContextLZ}). In this context, we can compute a policy that ensures success regardless of the values of $p_1$ and $p_2$. 
In other words, the contextual deployment face of safe landing locations in the urban context alleviates the need to test our model for many possible values of $p_1$ and $p_2$.
Specifically, the \textsc{uav}'s policy would have it fly between building rooftops only when it can safely cross without collision and loiter at the rooftop otherwise.
Of course, such a policy may result in extremely long loitering times while the \textsc{uav} waits for the delivery robot to move away from the goal region. 
We could thus make the design decision to include battery charge as an additional parameter in the \textsc{uav} parametric MDP system model.
This design decision could change the acceptable deployment contexts, though the details depend on what was learned through exploratory testing.

\section*{Certifying autonomous systems in sociotechnical contexts}
\label{sec:cert}

Testing in static certification can be tractable because the target performance is specified ahead of time~\cite{kaner:2003}. In contrast, testing in dynamic certification might appear completely intractable as it depends on the changing system, use, and context. We propose that the integration of modeling and testing can make dynamic certification feasible. A formal model can provide precise, context-sensitive specifications for the system's implementation and inform the types of tests we conduct. This type of dynamic certification will ideally result in \emph{believable} and \emph{defensible} guarantees of correct operation.\footnote{We cannot require infallible guarantees, as they may be based on incorrect or imperfect assumptions. No certification process can be perfect, but dynamic certification has the benefit of continued testing to detect incorrect (formal) models.} More importantly, this dynamic certification leads to early-phase models that can be used to interrogate required or acceptable behavior, even in the absence of a specific software or hardware implementation. Compared to conventional certification regimes, dynamic certification revises our assumptions and improves decisions or requirements before the system is even built, all with the added benefit of identifying the types of contexts that led to design changes. The effort to understand required assurances can begin while we can still effectively change the design or the broader sociotechnical context. 

Dynamic certification systematically identifies context-dependent stakeholder values and incorporates them in modeling and testing autonomous systems. We have outlined one way of implementing dynamic certification using formal methods and models. In contrast with the current practice of using formal methods for guarantees once a system is built, we can also use formal methods to model values and restrictions within deployment scenarios and open environments. Formal methods can simulate sociotechnical parameters, not just the technological system. The central message of dynamic certification is that we must implement precise feedback loops between formal models, simulation environments, and increasingly ``open world'' deployments, all to ensure that stakeholder values are being protected and advanced. Formal models provide a crucial tool in these loops as they can justify dynamically evolving requirements.

A common concern when implementing verification and certification using formal methods is scalability. However, scalability is not an issue within dynamic certification because we expect the formal model to provide partial proof of safe deployment. Dynamic certification augments formal models with testing for this reason. Indeed, the feedback between models, testing, and stakeholder values minimizes the scalability issues found in most static certification contexts. The latter must test all behaviors of systems, while the former can focus on the behaviors that lead to losses (violation of stakeholder values)~\cite{leveson:2011}. Formal methods thus do not need to scale indefinitely,\footnote{This does not mean that bottlenecks can not be hit with partial models, but partial modeling gives us the means to say something meaningful about the relationship between what the system ought (not) to do (i.e., its requirements) and what the exhibited behavior actually is.} does not mean that we could instead require regulators and designers to tame some of the environmental complexity or limit the required autonomy. Alternately, one could require only that the formal model demonstrate resilience in the sense of return to normalcy after an uncontrolled action. This narrower requirement can vastly reduce the scenarios for formal modeling and assurance~\cite{hosseini:2016}, as others can specify what is required for ``normal behavior.''

Dynamic certification differs from conventional certification not because it proposes stages and feedback loops---already present in static certification---but on the types of testing (exploratory, not just confirmatory) and specification (partial, rather than complete) in every stage. The more precise data we can capture with models and tools, the better-informed stakeholders will be to ensure the operational needs of the system. Toward this goal, research must be conducted at the intersection of robotics, control, learning, safety, security, resilience, testing, and formal methods. For example, roboticists must include realistic dynamical models for surrounding information that can be given by learning~\cite{sekhon:2021}, learning must be interpretable based on test vectors~\cite{sekhon:2019}, control must account for clashing safety requirements based on dynamics~\cite{leveson:1994}, and safety~\cite{lecomte:2019}, security~\cite{voas:2016}, and resilience~\cite{bouvier:2021} must be given formal interpretations that are based on realism but allow partial modeling, precisely to account for the uncertainty arising from coupled learning systems. Two recent improvements that will assist with developing dynamic certification are compositional verification, which relates different model types~\cite{bakirtzis:2021c}, and more operational data, e.g., high-definition maps for streets in major cities~\cite{argoverse:2022}.

Dynamic certification is an approach for autonomous systems that attempts to provide a common language between formal models, simulations, real-world (testing) data, and regulatory mechanisms. Dynamic certification requires advances in formalism compatibility and codesign, the development of high-fidelity simulation tools that can input information from formal models, expansive context-aware testing vectors, and legal codification of acceptable stages of deployment. In light of these multidisciplinary aspects, it is unsurprising that dynamic certification has been a relatively under-explored approach. However, dynamic certification promises better-designed, safer, and more secure autonomous systems, providing assurance of correct behavior and increased deployment of those systems. The effort to advance dynamic certification can provide significant benefits.

At the same time, AI presents additional challenges for the dynamic certification of autonomous systems. First, the distributed nature of much AI and robotic development can lead to significant communication barriers between different stakeholders during the requirements elicitation stage, and research is needed to develop, test, and validate structured approaches for requirement and value elicitation. Second, modular and scaleable methods and tools are needed to characterize precisely---whether through formal methods or otherwise---the connections between requirements and system (mis)behavior, particularly given the inevitable uncertainties with AI-enabled systems. Third, higher-fidelity causal models could enable improved counterfactual reasoning in the design and certification of autonomous systems, as the certification processes could then incorporate additional feedback loops that identify counterexamples in data collection, provide diagnostic capabilities, and clarify assumptions used to evaluate performance of the autonomous system in uncertain, ``open world'' environments.

\bibliographystyle{ACM-Reference-Format}
\bibliography{manuscript}
\end{document}